%% file: main.tex
\documentclass[10pt,twocolumn,letterpaper]{article}

\usepackage[pagenumbers]{cvpr}      

\usepackage{multirow}
\usepackage{colortbl}
\usepackage{makecell}
\usepackage{float}


\definecolor{cvprblue}{rgb}{0.21,0.49,0.74}
\usepackage[pagebackref,breaklinks,colorlinks,allcolors=cvprblue]{hyperref}

\def\paperID{15631} 
\def\confName{CVPR}
\def\confYear{2025}

\title{MaskGWM: A Generalizable Driving World Model with Video Mask Reconstruction}

\author{Jingcheng Ni, Yuxin Guo, Yichen Liu, Rui Chen, Lewei Lu, Zehuan Wu \\
SenseTime Research\\
Project page: \href{https://github.com/SenseTime-FVG/OpenDWM}{https://github.com/SenseTime-FVG/OpenDWM}
}

\newcommand{\ourmethod}{{MaskGWM}} 

\begin{document}
\maketitle
\input{sec/0_abstract}  
\input{sec/introduction} 
\input{sec/related_work} 
\input{sec/method}   
\input{sec/experiments}
\input{sec/conclusion}
{
    \small
    \bibliographystyle{ieeenat_fullname}
    \bibliography{main}
}
\input{sec/X_suppl}


\end{document}


\maketitle

1. Discussions
2. Implementation Details
3. More ablation Studies
4. More Results
{
    \small
    \bibliographystyle{ieeenat_fullname}
    \bibliography{main}
}


%% file: sec/0_abstract.tex
\begin{abstract}
World models that forecast environmental changes from actions are vital for autonomous driving models with strong generalization. 
The prevailing driving world model mainly build on video prediction model. Although these models can produce high-fidelity video sequences with advanced diffusion-based generator, they are constrained by their predictive duration and overall generalization capabilities. In this paper, we explore to solve this problem by combining generation loss with MAE-style feature-level context learning. In particular, we instantiate this target with three key design: (1) A more scalable Diffusion Transformer (DiT) structure trained with extra mask construction task. (2) we devise diffusion-related mask tokens to deal with the fuzzy relations between mask reconstruction and generative diffusion process. (3) we extend mask construction task to spatial-temporal domain by utilizing row-wise mask for shifted self-attention rather than masked self-attention in MAE. Then, we adopt a row-wise cross-view module to align with this mask design. 
Based on above improvement, we propose MaskGWM: a \textbf{G}eneralizable driving \textbf{W}orld \textbf{M}odel embodied with Video \textbf{Mask} reconstruction. 
Our model contains two variants: MaskGWM-long, focusing on long-horizon prediction, and MaskGWM-mview, dedicated to multi-view generation.
Comprehensive experiments on standard benchmarks validate the effectiveness of the proposed method, which contain normal validation of Nuscene dataset, long-horizon rollout of OpenDV-2K dataset and zero-shot validation of Waymo dataset. Quantitative metrics on these datasets show our method notably improving state-of-the-art driving world model.
\end{abstract}

%% file: sec/introduction.tex
\section{Introduction}
\label{sec:intro}

\begin{figure*}[t]
  \centering
  \begin{subfigure}{0.66\linewidth}
      \centering
      \begin{tabular}{@{}l|ccc|cc@{}}
        \toprule
        \multirow{2}{*}{\textbf{Method}} & & \textbf{Model Setup} & &  \\ 
        & Data scale & Framework & Multi-view & Traget\\
        \midrule
        Drive-WM ~\cite{drive-wm} & 5h & Unet & $\surd$ & Diff \\
        DiVE ~\cite{dive} & 5h & DiT & $\surd$ & Diff \\
        GenAD ~\cite{genad} & 1740h & Unet & $\times$ & Diff \\
        Vista ~\cite{vista} & 1740h & Unet & $\times$ & Diff \\
        \midrule
        \ourmethod (Ours) & 1740h & DiT & $\surd$ & Diff+MR \\
        \bottomrule
      \end{tabular}
      \caption{Real-world multi-view driving world models.}
      \label{tab:example}
  \end{subfigure}
  \hfill
  \begin{subfigure}{0.30\linewidth}
    \centering
    \includegraphics[width=1\linewidth]{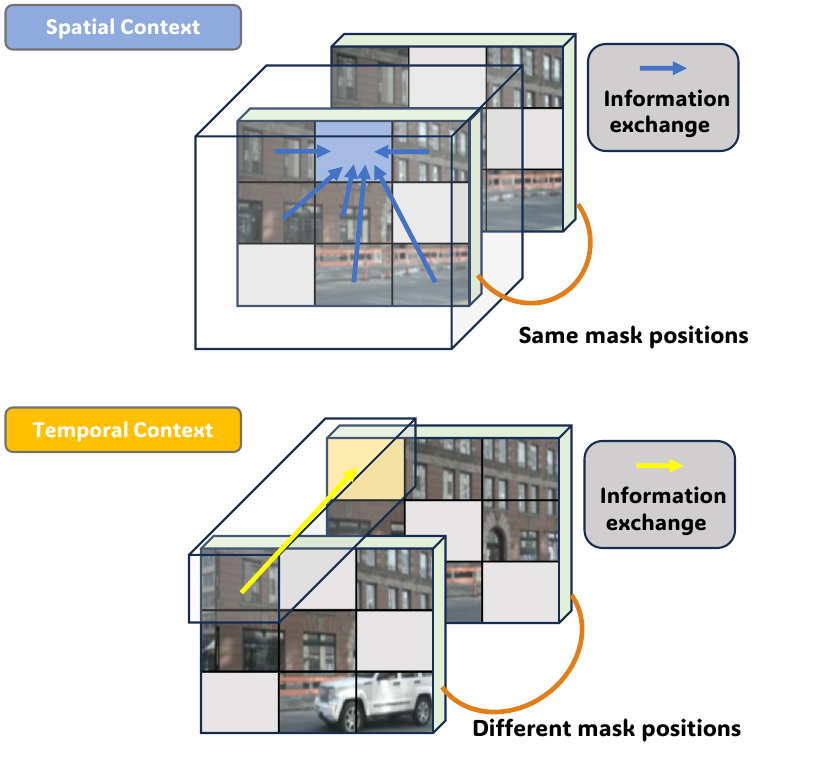}
    \caption{Different context for mask reconstruction.}
  \end{subfigure}
  \caption{(a). \ourmethod\ improve fidelity and generalization from web-scale dataset, scalable DiT architecture and  Mask Reconstruction (MR) target. (b) proposed MR apply a two branch structure for spatial context (scene objects) and temporal context (object motions)}
  \label{fig:short}
\end{figure*}

\begin{figure*}[h]
  \centering
   \includegraphics[width=1\linewidth]{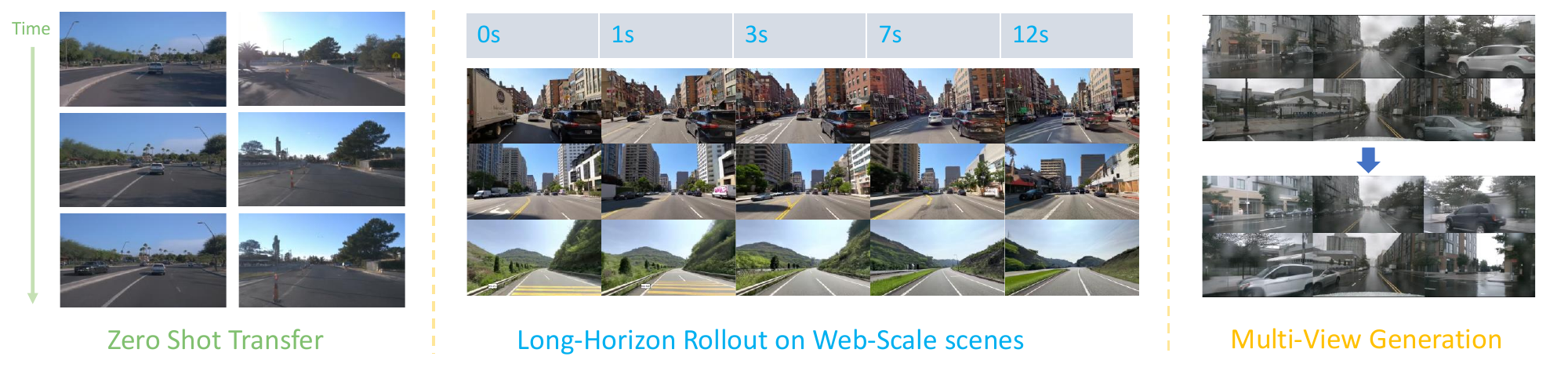}
   \caption{
   Our model facilitates  zero-shot generation, consistent long-horizon prediction and multi-view video generation.}
   \label{fig:teaser2}
\end{figure*}

As an pivotal application of artificial intelligence, autonomous driving technologies, which require comprehending the surrounding environment and executing correct actions, have achieved significant advancements following the emergence of various learning models~\cite{li2022bevformer,hu2023planning} with scalability. 
However, the challenge of limited generalization to complex and varied scenarios remains unresolved for state-of-the-art methods~\cite{li2024ego}.
For example, perception may encounter performance drops~\cite{xie2024benchmarking} in cases like weather changes, scene variations and motion blur.
A promising solution for this problem is the use of world models, which directly predict environment changes under different actions. These models facilitate unraveling the complexities of data distributions and craft intricate regular patterns like human perception system~\cite{lecun2022path}.

Recently, advanced methods~\cite{genad,vista,gaia1,drive-wm,drivedreamer,drivedreamer2,drivescape} developed world models through diffusion-based generation task, capitalizing on the rapid development of advanced image generation systems~\cite{sd3,sd,svd}. 
Despite generating high-fidelity results, these approaches still struggle with long-horizon prediction and zero-shot generalization. To address this, GenAD ~\cite{genad} attempts to conduct training on large-scale OpenDV-2K~\cite{genad} dataset with carefully-designed temporal modules, while VISTA~\cite{vista} further introduces explicit re-weighted generation loss on structural and moving areas. 
However, two problems still exist in building a generalizable world model for autonomous driving. First, the combination of large-scale training dataset with more scalable transformer architectures is still under exploration. Second, one fundamental question remains unanswered: Is diffusion-based generation sufficient to build a generalizable world model? Since diffusion loss targets at iterative de-noising, the learning of visual semantics may not be straightforward. For example, MaskDiT~\cite{maskdit} has shown diffusion models are complementary to well-known self-supervised methods~\cite{mae}, benefiting both convergence speed and generation quality.


Based on these insights into the diffusion pipeline and data for our driving world model, we design a new model, dubbed as MaskGWM, aiming at improving the fidelity, generalizability and long-time series prediction of the existing methods. Additionally, our model can also generate multi-view cases, by incorporating a multi-view module. We adopt Diffusion Transformer (DiT) as our backbone, which is more scalable and could take the information from a variety of datasets.
Moreover, we introduce the mask reconstruction as a complementary task for generation.
Several impactful works~\cite{mae,simmim} have demonstrated that masked autoencoder is a powerful self-supervised method for representation learning from large-scale data and it is also extended to some diffusion methods~\cite{maskdit,sd_dit,mdt} as an additional supervision to improve the models' performance. 
Additionally, the features obtained by self-supervised learning is more contextually meaningful~\cite{dino}, which can be used as an auxiliary supervision to further improve the generation quality~\cite{gaia1} 
However, integrating existing mask reconstruction for image generation into driving world models is not straightforward. There are still two questions to answer: (1) How can we enhance the synergy between diffusion model and mask reconstruction? Though the Mask reconstruction improves the contextual reasoning ability, this task contradicts with diffusion steps which include high noise ratio obscuring the feature details. (2) What kind of mask strategy should we use for video data? Different from image generation, the prediction of driving future requires an understanding of not only the objects within the scene but also their dynamic movements.


Therefore, we develop several special designs to address the aforementioned issues: (1) We make use of the mask tokens to improve the synergy between mask reconstruction and diffusion models. Specifically, we propose a diffusion-related mask tokens (Sec.\ref{sec:3.2}) 
to initialized the invisible patches after DiT encoder. This special mask token can balance the learning of global and local features.
(2) We design a novel two-branch mask reconstruction strategy. For spatial modeling, we use a mask shared across all frames and reconstruct invisible tokens via spatial transformer, this mask strategy is similar to some video mask modeling methods~\cite{videomae,videomae_v2}. For temporal modeling, we introduce a frame-specific mask and recover masked tokens via temporal transformer. Unlike the spatial branch, we directly link the unaligned tokens on temporal dimension after masking, which can be seen as a shift augmentation restricted by a new-proposed row-wise policy. We find this temporal branch achieves both masked patches prediction in the temporal context and a reduction in training costs.



In summary, our main contributions are:
\begin{itemize}
    \item We propose MaskGWM, a generalizable DiT-based driving world model capable of forecasting long-term futures across web-scale scenes.
    \item We introduce mask reconstruction as a complementary task for diffusion-based world model. 
    \item Comprehensive experiments on nuScene, OpenDV-2K and Waymo datasets demonstrate the superior video quality and robust generalization capabilities across extended time spans and different viewpoints.
\end{itemize}





%% file: sec/related_work.tex
\section{Related Works}
\label{sec:related}
\begin{figure*}[t]
  \centering
   \includegraphics[width=1\linewidth]{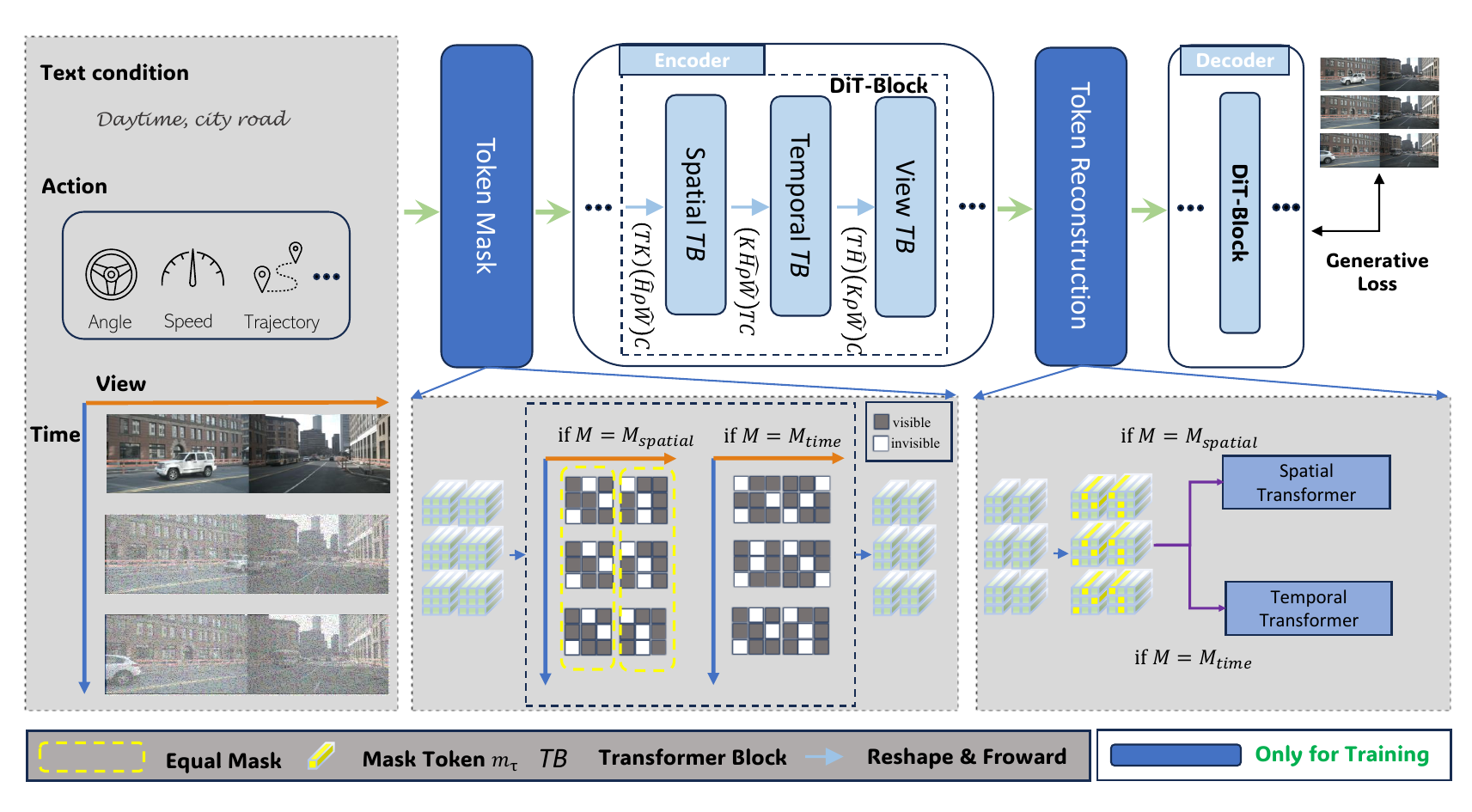}
   \caption{Overview of the \ourmethod. We propose mask reconstruction containing token mask and token reconstruction as a complementary task for training dring world model. \textbf{Token Mask}: we randomly sample tokens by temporal-shared $\mathcal{M}_{spatial}$ and temporal-unshared $\mathcal{M}_{time}$, specialized for spatial and temporal modeling. \textbf{Token Reconstruction}: we fill invisible tokens by diffusion-related mask tokens (Sec.\ref{sec:3.2}) and recover features by a two-branch transformer. Moreover, we introduce a row-wise mask strategy (Sec.\ref{sec:3.3}) for temporal branch. $\rho=1-r$ is used for simplicity in encoder.}
   \label{fig:framework}
\end{figure*}

\subsection{World Models}
World models aim to infer the dynamic environment and ego state from past observations for accurate future predictions and planning. Most studies achieve world understanding by enabling the model to generate realistic videos that align with physical principles. In autonomous driving area, world models primarily focus on generating controllable real-world driving scenarios. GAIA-1 \cite{gaia1} employs an autoregressive transformer to predict tokens based on past state and then leverages a diffusion decoder to generate high-quality videos. In contrast, DriveDreamer \cite{drivedreamer} directly use diffusion model to represent complex environments and generate driving videos from multi-modal inputs. Drive-WM \cite{drive-wm} extends to consistent and controllable multi-view video generation, exploring its application in end-to-end planning. GenAD \cite{genad} enhances generalization capability of world models by training on large-scale datasets and Vista~\cite{vista} achieves further improvements by introducing attention on structure and dynamic area.

\subsection{Diffusion Models with Self-supervised Learning}
Recently, diffusion-based methods~\cite{videodiffusionmodel, guo2023animatediff, henschel2024streamingt2v, latentdiffusion, controlnet} have become the mainstream of image and video generation. 
One important advancement in this field is Diffusion Transformer(DiT)~\cite{dit}. Due to its better scalability and lower computational cost, DiT has been successfully applied in various diffusion models, achieving state-of-the-art results~\cite{zheng2024viton, dive, sd3, feng2024dit4edit}. 

On the other hand, masking strategies from self-supervised learning have been effectively applied to enhance generative models. 
With the development of DiT, research has focused on migrating this self-supervised approach to diffusion-based models. 
Initial works like MDT~\cite{mdt} modify DiT blocks to an asymmetric masking diffusion transformer architecture, where the encoder handles unmasked tokens only and a side-interpolater is introduced to recover the latnet to the original shape. 
Following MDT~\cite{mdt}, MaskDiT~\cite{maskdit} 
simply utilizes a learnable token to fill in the masked places. 
SD-DiT~\cite{sd_dit}, noticing the training-inference discrepancy and fuzzy relations between mask strategy and diffusion process, introduce a novel masking DiT with self-supervised discrimination.
Despite their succuss, none of them applied the masking diffusion to the video generation models. Our work make this attempt on driving world model by applying different design of spatial context and temporal context, which focus on scene objects and nuance motion separately.

%% file: sec/method.tex
\section{Method}

Fig.~\ref{fig:framework} illustrates an overview of the proposed pipeline.
\ourmethod~builds upon Stable Diffusion 3 (SD3)~\cite{sd3}, which is a well-studied  DiT-based Text-to-Image (T2I) generation model, and introduce additional spatial and temporal blocks to extract the cross-view and cross-frame information. Moreover, we deploy a mask reconstruction module during training to improve the performance of our model. In this section, we first briefly review our DiT-based driving world model in Sec.~\ref{sec:3.1}. Then, Sec.~\ref{sec:3.2} details the pipeline for mask reconstruction and introduces novel diffusion-related mask tokens designed to enhance the synergy between diffusion generation and mask reconstruction. Following this, Sec.~\ref{sec:3.3} describes the extension of mask reconstruction to the temporal dimension. Finally, in Sec.~\ref{sec:3.4}, we describe the details of our cross-view module. 
\subsection{Preliminaries}
\label{sec:3.1}


\noindent\textbf{Diffusion Models.}
A mutli-view video sampled from a video dataset $p_{data}$ can be represented as $x_0 \sim p_{data}$, where $x_0 \in \mathrm{R}^{T\times K \times C \times H \times W}$ is a sequence of $T$ frames with view $K$, height $H$ and width $W$. We first transform $x_0$ into video tokens $z_0 = \mathrm{P}(\Theta(x_0)) \in \mathrm{R}^{T\times K \times C \times \hat{H} \times \hat{W}}$ via latent encoder $\Theta$ and patch encoding $\mathrm{P}$. \ourmethod\ applies Rectified Flow~\cite{sd3,liu2022flow,lipman2022flow} to model the generation process. Specifically, given a standard normal distribution $\epsilon \sim \mathcal{N}(0, \mathrm{I})$, Rectified Flow defines the intermediate noisy state as $z_\tau = (1-\tau)z_0 + \tau\epsilon$, where $\tau\in[0,1]$ is the diffusion timestep. The training target of Rectified Flow 
adopts the $v$-prediction, 
defined as: 
\begin{equation}
    \mathcal{L} = 
    \mathbb{E}_{{z_0},\epsilon\sim{\mathcal{N}(0, \mathrm{I})},\tau}\left[ \Vert
    G_\theta(z_\tau,\tau,c,\mathcal{M}) - (z_0-\epsilon)
    \Vert^2_2 \right],
\label{eq:loss}
\end{equation}
where c is the condition, $\mathcal{M}$ is a binary mask for mask reconstruction and $G_\theta$ is the DiT model. 

\noindent\textbf{Temporal Modeling.}
To facilitate temporal context learning, we attach a temporal transformer block after each 2D spatial transformer block, following common practice in video generation models~\cite{videoldm, Open-Sora}. During the forward process, to standardize the inputs to the different self-attention layers in the transformer blocks, we reshape the video latent representation to $\mathrm{(TK)(\hat{H}\hat{W})C}$ for spatial self-attention and to $\mathrm{(K\hat{H}\hat{W})TC}$ for temporal self-attention. Additionally, we introduce reference frames according to video DiT models~\cite{Open-Sora, gao2024lumina}. During diffusion process, diffusion timestep $\tau$ of the reference frames is always set as $0$, while the following frames to be predicted are embedded with regular timestep. 

\noindent\textbf{Unified Action Conditioning.}
Following VISTA, we provide nuanced control over low-level actions including angle, speed, trajectory and goal point, combining with high-level command capabilities. We construct action embedding by concatenating the Fourier embeddings~\cite{tancik2020fourier} of all actions. Subsequently, these action embeddings are projected and added to the key and value features of the cross-attention layers in temporal transformer blocks.

\subsection{Diffusion-related Mask Reconstruction}
\label{sec:3.2}
Motivated by previous works~\cite{maskdit,sd_dit}, Masked Image Modeling (MIM)~\cite{mae,simmim} with mask reconstruction object has been adopted to diffusion-based generation model and achieve improvement on training efficiency and local contextual perception. However, these methods fail to consider the influence of diffusion process, which incorporate noise schedule and complex training target. Therefore, we propose this novel pipeline integrated with mask reconstruction and then introduce our diffusion-related mask tokens for compatibility with diffusion process, which can reduce the effect of noise during diffusion process.
\begin{figure*}[t]
  \centering
   \includegraphics[width=1\linewidth]{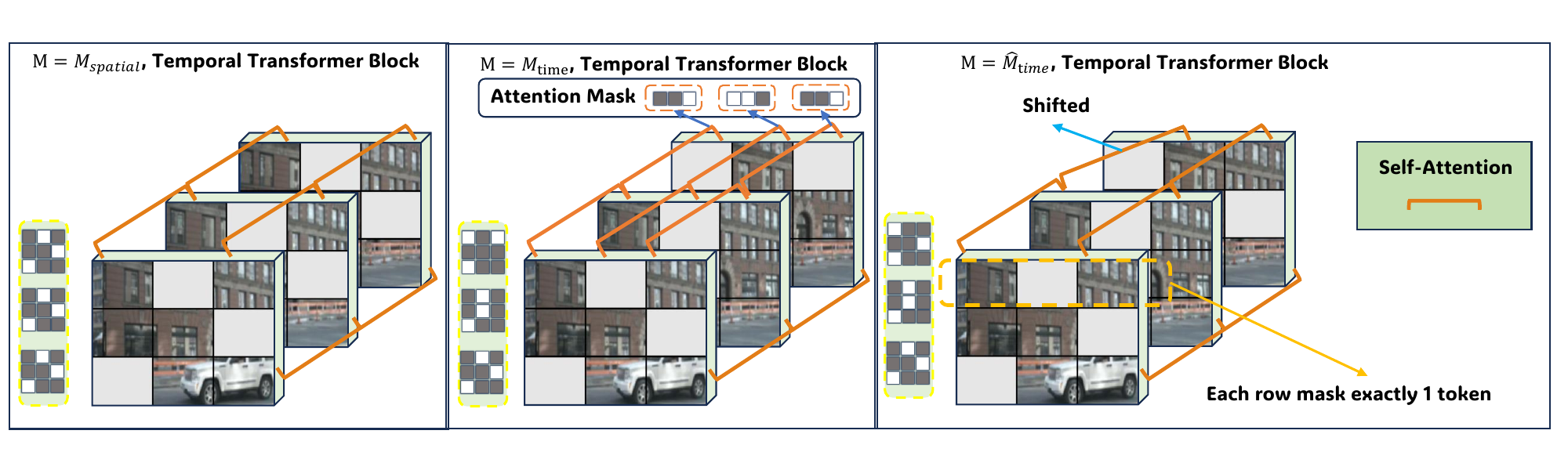}
   \caption{The comparison of different mask types and attention operations for temporal transformer block with mask reconstruction task. Attention mask is only applied when $\mathcal{M} = \mathcal{M}_{time}$}
   \label{fig:method2}
\end{figure*}

\noindent\textbf{Mask Reconstruction.} 
During the training phase, the DiT backbone is asymmetrically divided into  an encoder 
$E$ and a decoder $D$ for mask reconstruction and includes two additional processing steps. In the encoding stage, \ourmethod~produces a token mask by random sampling a binary mask $\mathcal{M} \in \mathrm{R}^{T \times K \times 1 \times H \times W}$, given the video latent $z_\tau$ at timestep $\tau$. Similar to~\cite{mdt,sd_dit}, $\mathcal{M}$ is only use to partition $z_\tau$ into visible patch tokens $z_\tau^v=z_\tau \odot \mathcal{M}$ and invisible patch tokens $z_\tau^{iv}=z_\tau \odot (1-\mathcal{M})$. 
In the decoding stage, an extra token reconstruction module is introduced to handle dropped invisible patches. Mask tokens $m_\tau$, representing invisible patches, are infilled at the positions of the dropped tokens. Then, a transformer block $F$ is utilized to provide contextual awareness from visible patches. In details,
\begin{equation}
    G_\theta(z_\tau,\mathcal{M}) = D(F(E(z_\tau^v)\odot \mathcal{M}+m_\tau \odot (1-\mathcal{M}))),
\label{eq:img_diff}
\end{equation}
where $\tau,c$ in Eq.\eqref{eq:loss} are ignored for simplicity. Note that mask reconstruction is skipped for inference in which all tokens are visible, equivalent to $G_\theta(z_\tau) = D(E(z_\tau)) = D(E(z_\tau^v + z_\tau^{iv}))$. In practice, invisible patches $z_\tau^{iv}$ are directly dropped during the encoding of training to enable memory and speed benefits.

\noindent\textbf{Diffusion-related Mask Tokens.}
SD-DiT~\cite{sd_dit} describes a fuzzy relationship between generation process and mask reconstruction. Concretely, mask reconstruction focuses on context reasoning while generation diffusion process aims to model the translations between real and fake distributions.
From the viewpoint of diffusion model, the mask reconstruction for represent learning can be regard as $z_0$-prediction task, whereas rectified flow employs $v$-prediction (pred $z_0-\epsilon$). Therefore, simply combining these two distinct objectives may not lead to good performance of the diffusion model, as also demonstrated by MaskDiT~\cite{maskdit}.

Existing works either instantiate mask tokens as learnable parameters~\cite{mae,mdt} $p$ or directly take noisy tokens $z_\tau$ as input~\cite{sd_dit}. As discussed above, these mask tokens cannot balance these two targets due to the absence of explicit information for acquiring $\epsilon$. 
We bridge this gap by introducing $f_m(\epsilon)$ into the mask token, where $f_m$ is a small network for encoding noise $\epsilon$. Given that $\epsilon$ is explicitly provided, it is easier to recover the original mask reconstruction target for representation learning within diffusion pipeline. 
Other than explicitly alignment for two prediction tasks, we further take $\tau$ into consideration. 
Overall, we define our mask token with learnable parameters $p$ as:
\begin{equation}
    m_\tau = (1-\tau)f_m(\epsilon) + \tau p.
\label{eq:img_diff}
\end{equation}
Based on our experiments in Section~\ref{sec:ablation}, we designed this mask token accordingly. When $\tau$ is large, the generation is performed at a high-noise level, and fine-grained image details are unknown. We hypothesize that the learnable parameters can estimate an average distribution under specific conditions (e.g., text) and assist in guiding the prediction direction when appearance details are lacking \cite{yue2024exploring}. Conversely, on a low-noise level, mask reconstruction encourages the model to be attentive to local details of visible patches. Therefore, the model can leverage the visible information to recover the patches filled with noise (by $f_m(\epsilon)$).

\subsection{Mask Reconstruction Strategy}
\label{sec:3.3}
\textbf{Temporal and Spatial Mask.}
Previous methods~\cite{videomae,videomae_v2} extend MIM to video domain by sharing the random mask across time, where the mask $\mathcal{M}=\mathcal{M}_{spatial}=[\mathcal{M}^1, \mathcal{M}^2..., \mathcal{M}^T]$ satisfy $\mathcal{M}^i = \mathcal{M}^j$ when $i\neq j$. Despite the effectiveness on understanding tasks in the aforementioned methods, this masking strategy may not be suitable for temporal modeling on driving video prediction.
To incorporate temporal learning, we introduce a temporal unshared mask $\mathcal{M}_{time}$ satisfy $\mathcal{M}^i \neq \mathcal{M}^j$ when $i\neq j$. Then, we specialize $\mathcal{M}_{spatial}$ for spatial modeling and consider MIM with $\mathcal{M}_{time}$ for temporal modeling. Then, we make tasks synergy by devising a two-branch transformer block: 
\begin{equation}
    F=
    \begin{cases}
        F_s&  \text{if}\  \mathcal{M} = \mathcal{M}_{spatial}\\
        F_t&  \text{if}\  \mathcal{M} = \mathcal{M}_{time}
    \end{cases},
\label{eq:img_diff}
\end{equation}
where $\mathcal{M} \in \{\mathcal{M}_{spatial}, \mathcal{M}_{time}\}$, $F_s$ is a spatial transformer block and $F_t$ is a temporal transformer block.

\noindent\textbf{Row-wise Approximation.} 
The most straightforward strategy for masking video data is to directly apply a random mask $\mathcal{M}_{\text{time}}$ on each frame like~\cite{videomae,videomae_v2}, as shown in the middle column of Fig.~\ref{fig:method2}. However, this design cause tokens on masked positions should be masked rather than directly dropped, since temporal self-attention require all entries have the same sequence length and apply an attention mask to control sparsity.
As a result, this masked temporal self-attention requires a 3D attention mask to skip masked tokens, which leads to additional computational cost and precludes the direct application of optimization operators like FlashAttention~\cite{dao2022flashattention}.
To address the aforementioned issues, we employ shifted temporal self-attention for $\mathcal{M}_{\text{time}}$. As illustrated in the right column of Fig.~\ref{fig:method2}, we randomly mask the same number of tokens for each row. Similar to the spatial branch, all invisible tokens are dropped, and the visible tokens are directly connected. Consequently, this operation can be regarded as a shift in temporal self-attention, which adheres to the core idea of Masked Reconstruction (MR): masked tokens are invisible and are predicted from context during the decoding stage.
Additionally, rearranging the visible tokens row by row ensures the relevance of information in the temporal attention block during the encoding stage. 
In experiments, we find this design improves not only the training speed but also the generation metrics, especially on larger mask ratio $r$. Since this row-wise shifting allows nearby tokens to fill in the blanks of masked tokens, We analysis this phenomenon by token filling makes all tokens on temporal axis are retained and minor shift can facilitate the temporal block in context reasoning.

To formulate this row-wise temporal mask, we define the mask ratio as $r$.
To generate the mask for a $\hat H \times \hat W$ image latent at frame $t$, we randomly generate $\hat H$ one-dimensional mask, each having $r\hat W$ zero values, and concatenate them to formulate the final mask for this image, denoted by $\widehat{\mathcal{M}}_{\text{time}}$. We use $\widehat{\mathcal{M}}_{\text{time}}$ to replace $\mathcal{M}_{\text{time}}$ in our training.

\subsection{Mask Reconstruction for Cross-View}
\label{sec:3.4}
Our method can be extended to generate multi-view driving videos by introducing view transformer blocks. We propose a cross-view row-wise self-attention mechanism, which concatenates video features horizontally and computes attention across multi-view features on the same row. Specifically, given a feature tensor of shape $ TK\hat{H}\hat{W} $, we apply masking, resulting in a tensor of shape $ TKC\hat{H} [(1 - r)\hat{W}] $. We then reshape it into size $ (T\hat{H})[K(1 - r)\hat{W}]C $. 

The key insights of the proposed cross-view row-wise attention are twofold: First, since the vertical context can be modeled by spatial transformer blocks, row-wise feature exchange provides sufficient receptive field to extract multi-view information. Second, the proposed row-wise masking for reconstruction can also be utilized as data augmentation, since $Kr\hat{W}$ tokens of each row are randomly dropped. Because the proposed masked modeling task focuses on spatio-temporal modeling, we do not apply mask reconstruction along the view dimension.


%% file: sec/experiments.tex
\section{Experiments}
\subsection{Setup}
\input{tab/main_rt2}
\input{tab/main_gen}
\begin{figure*}[t]
  \centering
   \includegraphics[width=1\linewidth]{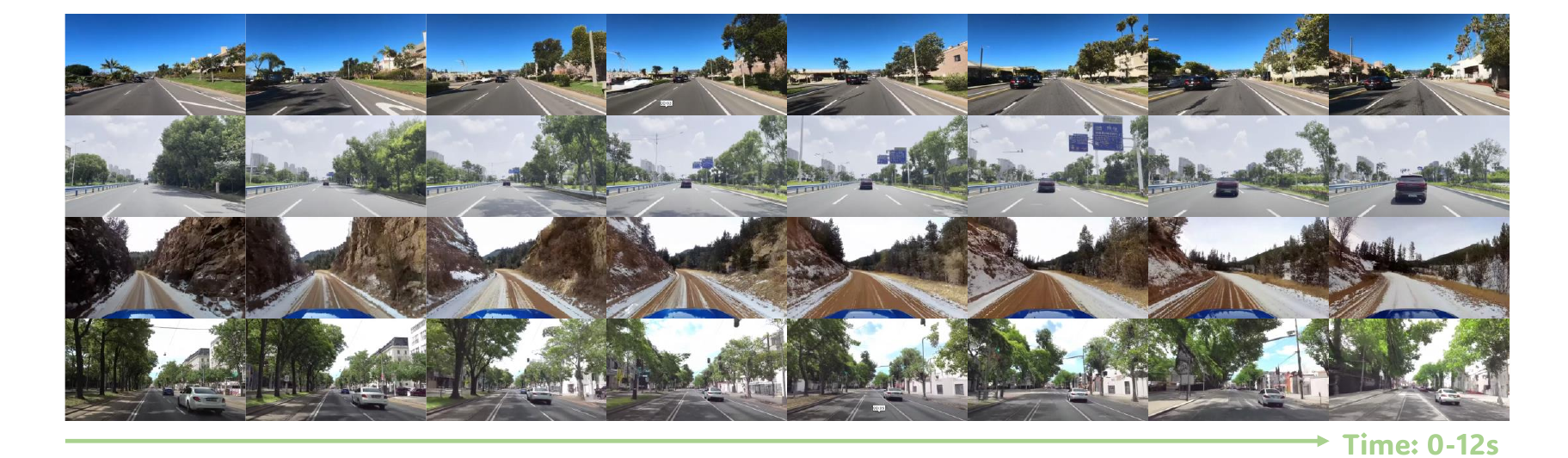}
   \caption{\textbf{Long-horizon prediction results of \ourmethod.}  Our model is capable of forecasting long video sequences with stability, devoid of collapse or blurring issues.}
   \label{fig:visual}
\end{figure*}
\begin{figure}[h]
  \centering
   \includegraphics[width=1\linewidth]{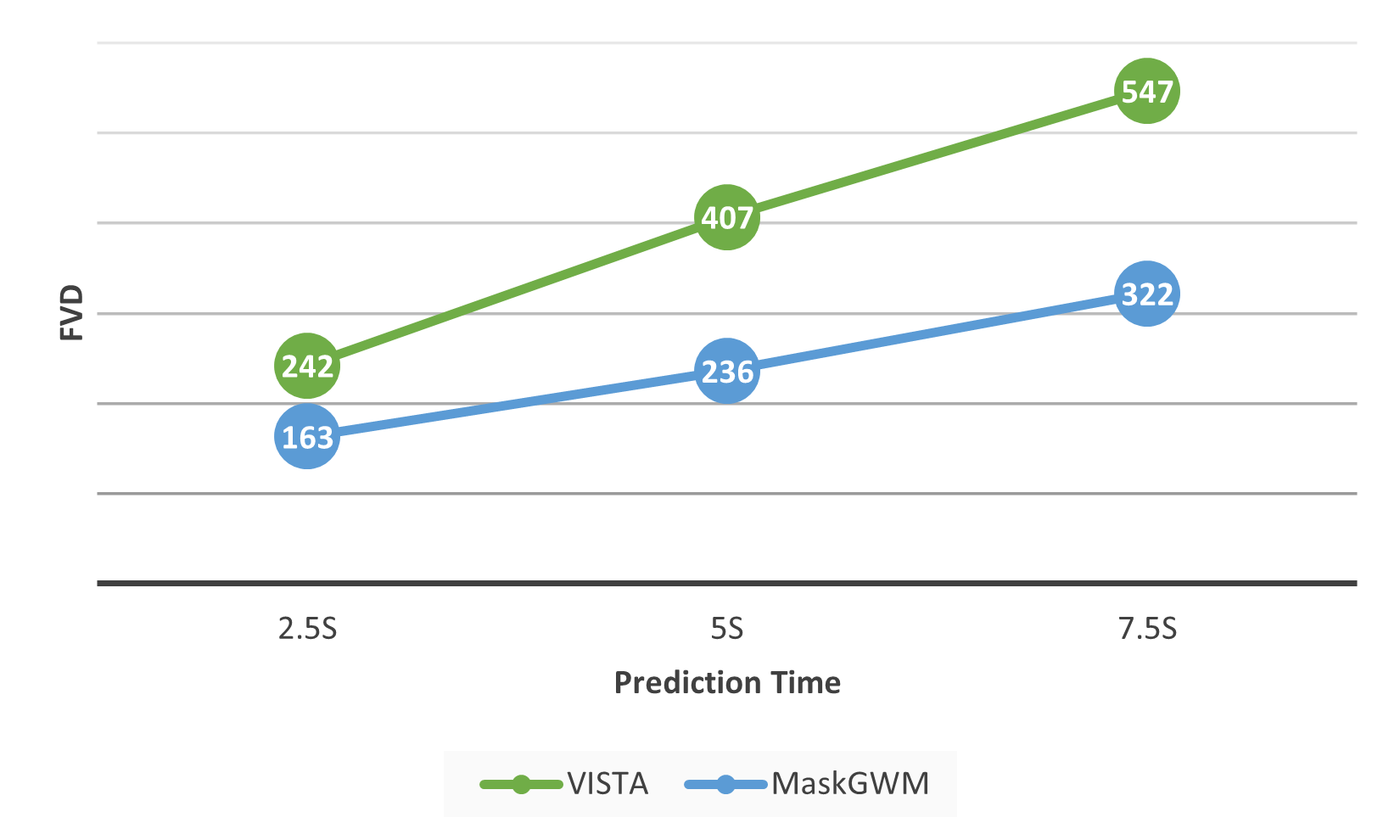}
   \caption{\textbf{Comparison of Long-horizon FVD metric on OpenDV2K validation set.} Our method demonstrates superior performance in terms of both value and growth rate.}
   \label{fig:long}
\end{figure}
\textbf{Datasets.} We conduct comprehensive experiments on a single-view dataset OpenDV-2K \cite{genad} and two multi-views datasets nuScenes \cite{caesar2020nuscenes} and Waymo \cite{waymo}. We follow the official splits to divide the training and validation sets.

\noindent \textbf{Evaluation.}
The quality of the generated images and videos are assessed using the Frechet Inception Distance (FID)~\cite{fid} for images and the Frechet Video Distance (FVD)~\cite{fvd} for videos. For fair comparison with previous works, we apply different evaluation settings for single-view model and multi-view model. For single view model, we align the evaluation setting of VISTA~\cite{vista}.
For multi-view model, we adopt the setting of Drive-WM~\cite{drive-wm}. Please refer to Appendix.\ref{sec:supp_sample_details} for more details.
To assess generalization ability, we evaluate zero-shot performance on Waymo validation set, using 600 videos for FVD and 15K frames for FID. 

\subsection{Training Scheme}
Our model is initialized with SD3~\cite{sd3} medium checkpoint with 2B parameters. 
There are three stages for our training: Stage 1 for pre-training on large-scale OpenDV-2K dataset, Stage 2 for single-view model \ourmethod-long and Stage 3 for multi-view model \ourmethod-mview. For all stages, we resize the original images to 512$\times$288, masking ratio $r$ is set to 0.25.

\noindent \textbf{Stage 1.} 
Following VISTA~\cite{vista}, we first pre-train our model on OpenDV-2K dataset. As our model starts from image backbone, we first train our model using single frame videos with batch size 768 for 18K iterations and then we train our temporal blocks with 16/20/24 frame videos and batch size 64. In particular, temporal blocks is initialized with zero and the training of temporal blocks takes 24K iterations in this step. Afterwards, we insert the zero-initialized reconstruction blocks $F$ and train with masking strategy for extra 20K iterations. The reason for varying frame length during training is that the frame numbers for single-view and multi-view models are different.

\noindent \textbf{Stage 2 for \ourmethod-long.} 
We devised \ourmethod-long to align with VISTA, where the frame length $T$ is set to 25 and cross-view blocks are skipped. We also follow the stage 2 of VISTA's collaborative training, in which data were sampled equally from nuScene and OpenDV-2K, and the action module is zero-initialized and trained.

\noindent \textbf{Stage 3 for \ourmethod-mview.}
Based on the well-trained \ourmethod-long, we enable multi-view ability of \ourmethod-mview by adding the cross-view blocks and train it on nuScene dataset for 6K steps and frame length $T$ is reduced to 8.


\input{tab/abl_mr}
\input{tab/abl_cv}
\subsection{Comparison}
\noindent\textbf{Generation Quality.}
Tab.\ref{tab:video-main} presents the quantitative comparison. In single-view generation, we achieve a remarkable FID of 5.6 and an FVD of 92.5, surpassing the previous state-of-the-art approaches~\cite{genad,vista} that are also trained on web-scale dataset. It is worth noting that there is a discrepancy between the training and evaluation phases in Vista, which integrates action during training but excludes it during evaluation. To address this, we also experiment with an aligned setting that drops action information for both training and evaluation. As indicated in Table \ref{tab:video-main}, this aligned setting yields significantly improved results, with a FID of 4.0 and a FVD of 59.4. We present these results for reference. For multi-view cases, similar improvement can also be observed, with 8.9 FID and 65.4 FVD. It is worth noting that our model directly predict multi-view future without the requirement of future layouts, unlike those methods~\cite{dive} deviating from video prediction task. Furthermore, our method represents a pioneering effort in extending a generalizable single-view model, trained on OpenDV-2K, to the domain of multi-view models.

\noindent \textbf{Generalization ability.}
In Tab.\ref{tab:main_gen}, we assess the generalization capability of our approach on the Waymo dataset, which is excluded from our training datasets. We conducted inference using the official checkpoint of VISTA~\cite{vista}. The results indicate that our method attains superior FVD while maintaining comparable FID, thereby demonstrating the generalization ability of our method.

\noindent \textbf{Long-horizon prediction.}
Fig.\ref{fig:long} illustrates the comparison of long-time prediction with VISTA. Due to the computational cost of generating long videos, both FID and FVD metrics are calculated on 300 videos randomly sampled from the OpenDV-2K validation set. The slope of FVD curve is significantly lower than VISTA, showing the less degradation. Qualitative results are illustrated in Fig.\ref{fig:visual} and appendix.





\subsection{Ablation Studies}
\label{sec:ablation}
We conduct comprehensive ablation study to verfiy the performance of every component in MaskGWM. Following the setting of GenAD for effective experiments, most ablation studies are conducted on stage 1 after mask reconstruction is inserted and the metrics are reported on the validation set of OpenDV-2K dataset with 3000 video clips for FVD and 18000 frames for FID. For the ablation of cross-view transformer blocks, we use multi-view nuScene metrics.

\noindent \textbf{Effect of mask tokens.} As indicated in Tab.\ref{tab:abl-1}, we make comparisons across different designs for mask tokens $m_\tau$. To analysis the influence of diffusion timestep $\tau$, we test only applying mask reconstruction on certain timestep range. We find learnable parameters $p$ shown better results on high noisy level and proposed noise embedding $f_m(\epsilon)$ perform better on low noisy level. This demonstrates the merit of noise embedding, which can recover the original mask reconstruction target for represent learning by explicitly giving $\epsilon$, achieving better result on generation steps with local details. In addition, we also try the mask tokens used in SD-DiT~\cite{sd_dit}, combining with extra contrastive~\cite{caron2021emerging} loss. The experiment shows our proposed diffusion-related mask tokens achieve best performance via combining the advantage of $p$ and $f_m(\epsilon)$.

\noindent \textbf{Effect of mask reconstruction.}
In Tab.\ref{tab:abl-2} and Tab.\ref{tab:abl-3}, we explore the effectiveness of proposed mask reconstruction in spatial-temporal domain. According to Tab.\ref{tab:abl-3}, we find that our row-wise time mask with shift attention mechanism assists a lot the model's convergence, while larger mask ratio can also make positive effect on the results. Thus, this setting is adopted in the ablation study on the mask strategy $\mathcal{M}$. As shown in Tab.\ref{tab:abl-2}, combining $\mathcal{M}_{spatial}$ and $\mathcal{M}_{time}$ can 
significantly improve the generation quality of the whole videos and each single images. We hypothesis this is because the temporal modeling is more sensitive to the dropout ratio than spatial counterpart (Please see more details on Appendix.\ref{sec:supp_abl_mr}). 
When shifted temporal self-attention is applied with a row-wise mask, all tokens on temporal axis are retained and experience only minor spatial shifts. Whereas invisible tokens are skipped without shift temporal self-attention, which results in a discrepancy between the number of tokens used in training and inference.
Moreover, the training speed is improved since invisible tokens dropped. From Tab.\ref{tab:abl-2}, we can see our two-branch mask reconstruction achieve best results, showing the effectiveness of introducing mask reconstruction on temporal context.

\noindent \textbf{Effect of cross-view module.}
As shown in Tab.\ref{tab:abl-5}, we make comparisons across different types of cross-view modeling. We find that introducing mask reconstruction on stage-3 training also yields favorable results. 
This indicates that randomly masking certain tokens along the view-row dimension is not detrimental and can even enhance the final results. This is different from temporal counterpart that requires shift-attention to avoiding tokens dropping. 

Furthermore, the experiment shows that proposed attention on dimensions $\mathrm{KW}$ outperforms view-attention on $\mathrm{K}$ used in previous methods~\cite{drive-wm,drivedreamer}. For view attention on $\mathrm{KHW}$, despite the minor improvement, the computation complexity explodes significantly. Consequently, this design is not adopted in our experiments.

\noindent \textbf{Effect of two-branch token reconstruction.}
We validate the effectiveness of two-branch transformer reconstruction structure for token reconstruction on Tab.\ref{tab:abl-4}. We remove two-branch structure by applying sequential spatial-temporal transformer blocks, which is shared by $F_s$ and $F_t$. 
We find the two-branch structure produces better results, especially on FVD. Unshared design forces the model to reconstruct masked features by corresponding context, leading to better spatial and temporal modeling for different conditions. 

%% file: tab/main_rt2.tex
\begin{table}[t]
  \small
  \centering
  \begin{tabular}{c|cccc}
  \toprule
  \textbf{Method} & \textbf{\makecell[c]{Multi-\\view}} & \textbf{\makecell[c]{Future-\\layout}} & \textbf{FID$\downarrow$} & \textbf{FVD$\downarrow$} \\
  \midrule
  DriveDreamer~\cite{drivedreamer} & $\surd$ & $\times$ &  14.9 & 340.8 \\
  MagicDrive~\cite{magicdrive} & $\surd$ & $\surd$ & 19.1 & 218.1 \\
  DiVE~\cite{dive} & $\surd$ & $\surd$ & - & 94.6 \\
  DriveDreamer-2~\cite{drivedreamer2} & $\surd$ & $\times$  & 11.2 & 55.7 \\ 
  Drive-WM~\cite{drive-wm} & $\surd$ & $\times$ & 15.8 & 122.7 \\
  \midrule
  \rowcolor{gray!20}
  \ourmethod-mview & $\surd$ & $\times$ & 8.9 & 65.4\\ 
  \midrule
  DriveGAN~\cite{drivegan} & $\times$ & $\times$ & 73.4 & 502.3 \\
  GenAD~\cite{genad} & $\times$ & $\times$ & 15.4 & 184.0 \\
  Vista~\cite{vista} & $\times$ & $\times$ & 6.9 & 89.4 \\
  \midrule
  \rowcolor{gray!50}
  \ourmethod-long & $\times$ & $\times$ & 5.6 & 92.5 \\ 
  \rowcolor{gray!50}
  \ourmethod-long$^\dagger$ & $\times$ & $\times$ & 4.0 & 59.4 \\ 
  \bottomrule
  \end{tabular}
  \caption{Performance comparison with state-of-the-art methods on nuScene Dataset. The varying shades of gray indicate our multi-view metric following Drive-WM and single-view metric following Vista for more fair comparison. Future layout refers to the availability of layout information for future time steps. $^\dagger$ denotes training without action.}
  \label{tab:video-main}
\end{table}

%% file: tab/main_gen.tex
\begin{table}[t]
    \centering
    \begin{tabular}{l|cc}
        \toprule
        \textbf{Method} & \textbf{FVD$\downarrow$} &       \textbf{FID$\downarrow$} \\
        \midrule
        VISTA~\cite{vista}$^\dagger$    & 176.56 & 9.76 \\
        \midrule
        \ourmethod-long    & 118.83  & 9.55  \\
        \bottomrule
    \end{tabular}
    \caption{Zero-shot metrics on 600 Waymo validation samples. $^\dagger$ denotes inference by official checkpoint.}
\label{tab:main_gen}
\end{table}

%% file: tab/abl_mr.tex
\begin{table*}[t]
    \centering
    \begin{subtable}[b]{0.42\linewidth}
    \begin{tabular}{ccc|cc}
        \toprule
        \textbf{$m_\tau$} & $\tau$ range & contra. &       \textbf{FVD$\downarrow$} &\textbf{FID$\downarrow$} \\
        \midrule
         $p$ & $[0, 1]$ &  & 126.71 & 7.35 \\
         $p$ & $[0.5, 1]$ &  & 120.35 & 7.12 \\
         $f_m(\epsilon)$ & $[0, 1]$ &  & 116.85 & 6.40 \\
         $f_m(\epsilon)$ & $[0, 0.5]$ &  & 109.87 & 5.92 \\
         $z_\tau$ & $[0, 1]$ & $\surd$ & 113.26 & 6.32 \\
         \midrule
         ours & $[0, 1]$ &  & 105.52 & 5.69 \\
        \bottomrule
        \end{tabular}
        \caption{Different design of $m_\tau$.}
        \label{tab:abl-1}
    \end{subtable}
    \begin{subtable}[b]{0.55\linewidth}
    \begin{tabular}{ccccc|cc}
        \toprule
        $\mathcal{M}_{time}$ & row & shift att. & $\mathcal{M}_{spatial}$ & $r$ & \textbf{FVD$\downarrow$} & \textbf{FID$\downarrow$} \\
        \midrule
        & & & & 0 & 136.55 & 10.28 \\
        $\surd$ &  & & & 0.25 & 142.68 & 10.75 \\
        $\surd$ & $\surd$ & & & 0.25 & 143.07 & 10.39 \\
        $\surd$ & $\surd$ & $\surd$ & & 0.25 & 121.38 & 7.36 \\
        & & & $\surd$ & 0.25 & 116.73 & 5.92 \\
        \midrule
        $\surd$ & $\surd$ & $\surd$ & $\surd$ & 0.25 & 105.52 & 5.69 \\

        \bottomrule
        \end{tabular}
        \caption{Ablations of two-branch mask reconstruction.}
        \label{tab:abl-2}
    \end{subtable}
    \caption{Ablations of the components in our mask reconstruction. ``contra'' stands for contrastive loss, ``row'' denotes whether $\mathrm{M}_{time}$ satisfies above row-wise generation strategy ($\mathrm{M}_{time}=\mathrm{\hat{M}}_{time}$) and "att." refers to self-attention }
\end{table*}

%% file: tab/abl_cv.tex
\begin{table*}[h]
    \centering
    \footnotesize
    \begin{subtable}[b]{0.33\linewidth}
    \begin{tabular}{cc|cc}
        \toprule
         row\&shift att & $r$ & \textbf{FVD$\downarrow$} & \textbf{Time$\downarrow$} \\
        \midrule
        & 0.1 & 133.24 & 0.368d \\
        & 0.25 & 142.68 & 0.352d \\
        $\surd$ & 0.1 & 123.70 & 0.357d \\
        $\surd$ & 0.25 & 121.38 & 0.329d \\

        \bottomrule
        \end{tabular}
        \caption{Different mask ratio $r$.}
        \label{tab:abl-3}
    \end{subtable}
    \begin{subtable}[b]{0.32\linewidth}
    \begin{tabular}{cc|cc}
        \toprule
        two-branch  & layers & \textbf{FVD$\downarrow$} & \textbf{FID$\downarrow$}  \\
        \midrule
         & 1 & 121.37 & 5.85 \\
        $\surd$ & 1 & 105.52 & 5.69 \\
         & 2 & 127.91 & 6.03 \\
        $\surd$ & 2 & 107.34 & 5.60 \\
        \bottomrule
        \end{tabular}
        \caption{Different design of $F$.}
        \label{tab:abl-4}
    \end{subtable}
    \begin{subtable}[b]{0.29\linewidth}
    \begin{tabular}{cc|cc}
        \toprule
        att dim & ratio $r$ & \textbf{FVD$\downarrow$} & \textbf{FID$\downarrow$} \\
        \midrule
        $\mathrm{KW}$ & 0 & 65.9 & 9.2 \\
        $\mathrm{KW}$ & 0.25 & 65.4 & 8.9 \\
        $\mathrm{K}$ & 0.25 & 71.5 & 8.7 \\
        $\mathrm{KHW}$ & 0.25 & 64.7 & 8.5 \\
        \bottomrule
        \end{tabular}
        \caption{Ablation of cross-view block.}
        \label{tab:abl-5}
    \end{subtable}
    \caption{Ablations of the impact of mask ratio $r$, different view transformer block designs and the effect  of mask reconstruction on convergence. "share" stands for applying one shared model for $F_s$ and $F_t$. \textbf{Time} indicate the training time for 10k steps.}
\end{table*}

%% file: sec/conclusion.tex
\section{Conclusion}
\label{sec:conclusion}
We introduce MaskGWM, the first DiT-based driving world model trained on web-scale datasets with masking strategy during training. By introducing a novel video dual-branch mask reconstruction, our model excels in both numeric metrics on fidelity and generation ability. Additionally, our mask policy also accelerate the training process and read memory consumption. Our extensive experiments showcase \ourmethod\  achieves the state-of-the-art performance on generation quality on nuScene, zero-shot ability on Waymo and long-time prediction ability on OpenDV-2K. These results furtehr indicates that our method can serve as a greate training programs to enable long driving video prediction.

%% file: sec/X_suppl.tex
\clearpage
\setcounter{page}{1}
\maketitlesupplementary

\input{tab/supp_abl_long}
\section{More Ablation Experiments}
\subsection{Additional Results of Mask Ratio}
\label{sec:supp_abl_mr}
Table~\ref{tab:supp_abl_mr} illustrates the impact of the mask ratio $r$ on mask reconstruction across various branches and temporal attention strategies. Our findings reveal several key insights: (1) The temporal branch equipped with masked temporal self-attention is more sensitive to mask ratio and necessitates a substantially lower mask ratio compared to the spatial branch. (2) The influence of the mask ratio on the proposed shifted temporal self-attention is more consistent with that observed on the spatial branch. As depicted in Fig.\ref{fig:method2}, the main difference in the DiT Encoder with the spatial branch is the positional shift, which can be effectively handled by positional encoding. Consequently, this allows for the attainment of an well-performed mask ratio (e.g. 0.25) in both spatial MR and temporal MR.

\input{tab/supp_abl_mr}
\subsection{Additional Results of Mask Reconstruction on NuScene Dataset}
As described in GenAD~\cite{genad}, the training and validation sets of OpenDV-2K are sourced from different YouTube videos with significant scene changes. Therefore, the model's performance on this dataset can be used for the evaluation of its generalization ability. We also conduct ablation studies in the in-domain setting by evaluating metrics on the nuScenes validation dataset. As shown in Table~\ref{tab:supp_abl_ns}, the proposed mask reconstruction method achieves significant improvements on both metrics.

\input{tab/supp_abl_ns}
\subsection{Additional Results of Mask Reconstruction on Long-Horizon Prediction}
To further analyse the influence of MR on auto-regressive generation, we extend the video duration to approximately 12 seconds and documented the metrics in Fig.\ref{fig:supp_abl_long}. The results indicate that MR is also effective in enhancing performance in long-sequence prediction. 
Although our baseline without MR still outperforms Vista, the quality of generation begins to deteriorate notably from about 8 seconds, and the FID score increases to 37.7 at the 10 second, making it also incapable to predict the distant future. Consequently, we conclude that this baseline’s improvements cannot translate to significant advancements in long-sequence.
In contrast, when MR is integrated into our method, the fundamental enhancements in single-step generation lead to significantly alleviate quality degradation over time. As a result, \ourmethod\ is capable of generating 10 Hz videos with discernible scene elements for a long time, and even 60 second examples, which far exceeds both Vista and our non-MR baseline. Therefore, we regard MR as a pivotal design that enables the model to make generalized predictions over extended durations. Note that this evaluation is conducted on 300 videos of OpenDV-2K validation set only, due to their longer video sequence. Thus, the single-step (2.5 seconds) FID and FVD are higher than results in Tab.\ref{tab:supp_abl_ns}, which is computed on 1800 videos.
\section{Implementation Details}
\subsection{Concrete DiT Structure}
We adopt the framework of SD3 and start our model with 2B parameters. We make several modifications to the original spatial transformer block to facilitate temporal and cross-view context modeling. 
First, Due to the limited availability of high-quality text data in our training datasets, e.g. only scene-level descriptions on nuScene, we skip the update of text feature by new-initialized temporal and cross-view transformer blocks. Then, for temporal transformer block, we make another modification to accommodate condition frames. To streamline the explanation, we represent the transformation within a transformer block as $z^\prime_{out} = z^\prime_{in} + f_b(z^\prime_{in})$, where $z^\prime_{in}$ and $z^\prime_{out}$ are the input and out features, respectively, and $f_b$ is the transformer block. Given the frame-level binary indicator $m_c$ with value 0 on condition frames, the diffusion time-step $\tau$, and time-step aware embeddings for scale $f_{scale}$ and shift $f_{shift}$, we introduce condition frames by:
\begin{equation}
    z^\prime_{out} = z^\prime_{in} + f_b(f_{scale}(m_c \tau)z^\prime_{in} + f_{shift}(m_c \tau))
\end{equation}
Here $m_c \tau$ is employed to reset the time-step for conditional frames to zero and time-step aware embeddings is applied for linear transform. 

We append one temporal transformer block and one view transformer block after each spatial transformer block following the common practice of previous works~\cite{drive-wm,vista}.

\subsection{Detailed Training Parameters}
We employ the Adam optimizer~\cite{loshchilov2017decoupled} for model training, using a learning rate of 5e-5. Throughout all training stages, we initiate the process with 1K warm-up steps and then maintain a constant learning rate. For condition frames, we randomly sample from zero to three frames following VISTA. We train Stage 1 for total 62K steps, Stage 2 for total 20K steps and Stage 3 for 6K steps. We select the training step based on numerical metrics from videos that are randomly sampled from the training set. Our training are conducted on 32 A800 GPUs with around 3 days on Stage 1.

\subsection{Detailed Sampling Parameters}
\label{sec:supp_sample_details}
Our sampling strategy does not incorporate any special designs. We generate the video by sampling 30 steps and utilize a classifier-free guidance scale~\cite{ho2022classifier} of 4.0.
Following Vista, we generate 25-frame videos containing one reference frame on full nuScene validation set with 5369 samples for our single-view model. All generated videos and corresponding frames are used for computing FVD and FID respectively. For our multi-view model, we generate 150 6-view videos for each nuScene scene, resulting in 900 single-view videos. Then,  10K frames are randomly sampled from these 900 videos for computing FID. This is align with the evaluation setting of DriveWM~\cite{drive-wm}.

\subsection{Details of Comparisons with Vista}
For comparisons with Vista, we use the official sample script and checkpoint. For zero-infer on Waymo dataset, we infer both models without action and the number of condition frame is set to 1. For long-horizon rollout on OpenDV-2K dataset, we infer both models without action and the number of condition frame is set to 3 for better temporal continuity across auto-regressive steps. We find numeric improvement is similar for one condition frame but the qualitative continuity is reduced by one-frame auto-regression. For auto-regressive steps larger than 1, we randomly select 25 frames from the generated video sequences to calculate the FVD and FID metrics.

\section{Qualitative Results}
\label{sec:rationale}
\subsection{Long-horizon rollouts}
\begin{figure*}[t]
  \centering
   \includegraphics[width=1\linewidth]{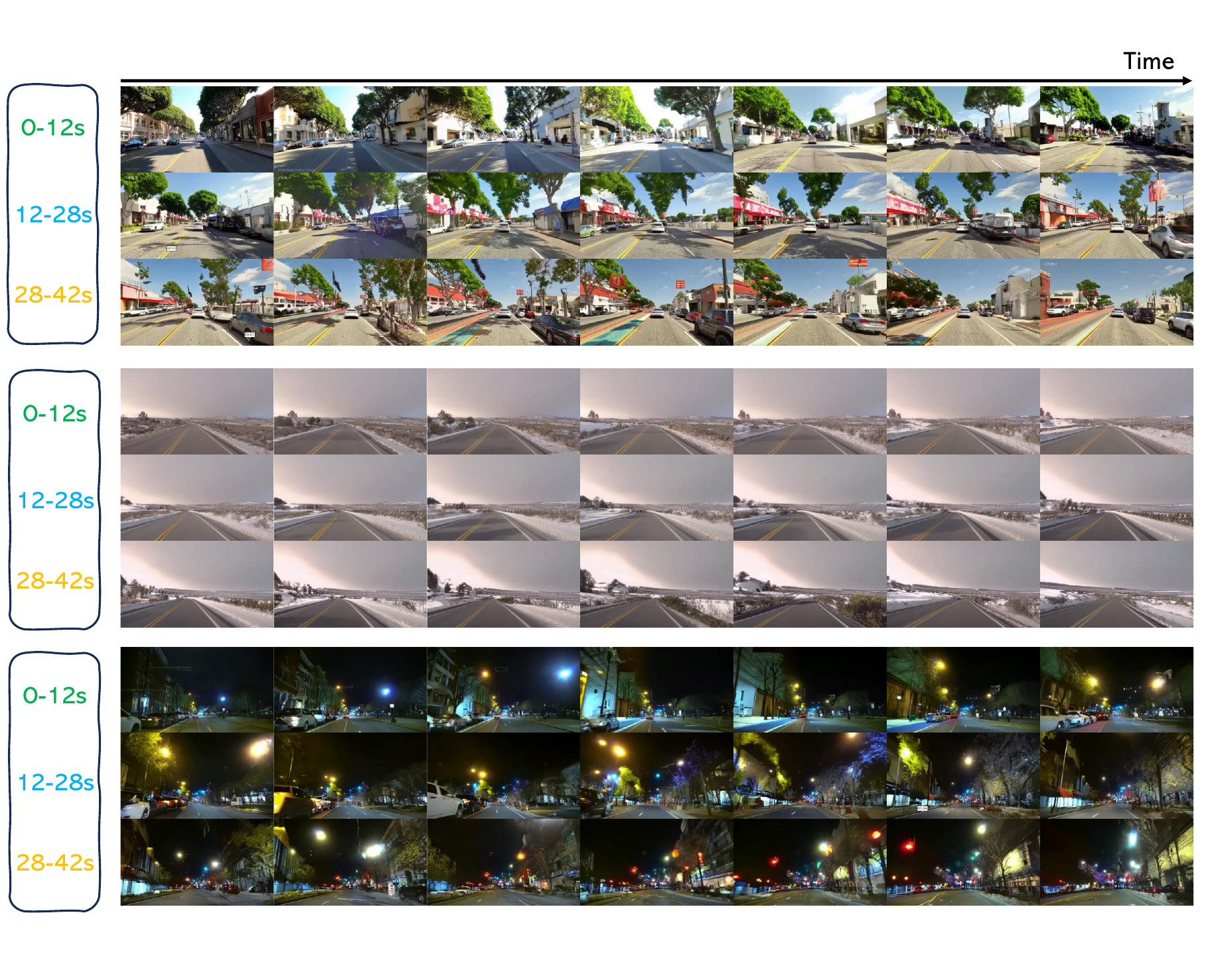}
   \caption{Generalization ability of \ourmethod\ with longer time.}
   \label{fig:vis_long2}
\end{figure*}

(what is rollout)
\noindent\textbf{Longer prediction} We provide more qualitative and longer visualizations with 42-seconds videos in Fig.\ref{fig:vis_long2}. We find \ourmethod\ can predict stable and consistent driving future, combined with unseen scene with initial scope.

\begin{figure*}[t]
  \centering
   \includegraphics[width=1\linewidth]{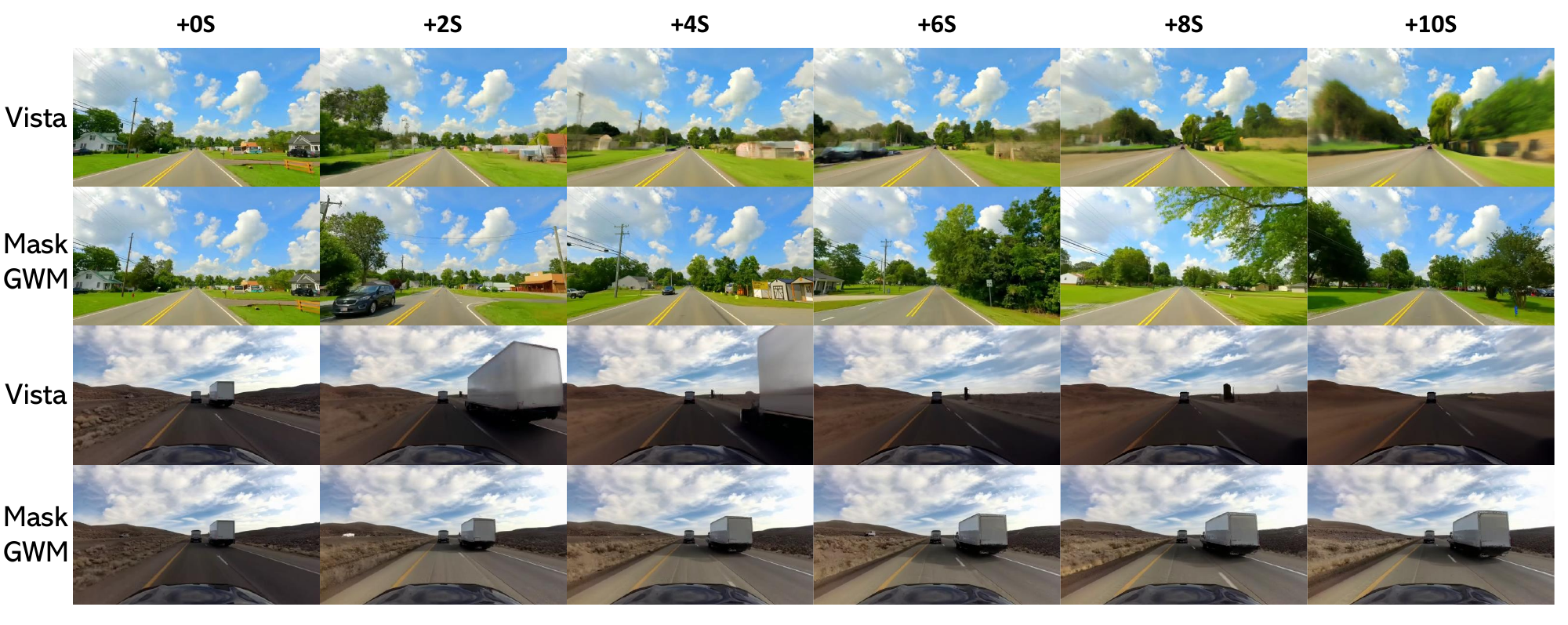}
   \caption{Qualitative comparison with Vista.}
   \label{fig:vis_cmp}
\end{figure*}

\noindent\textbf{Qualitative comparisons} In Fig.\ref{fig:vis_cmp}, we make qualitative comparisons with Vista, which is previous state-of-the-art method on generalizable driving world model. Our method can both make stable prediction and generate dynamic objects according to the future, e.g. unseen cars in initial visual scope.

\begin{figure*}[t]
  \centering
   \includegraphics[width=1\linewidth]{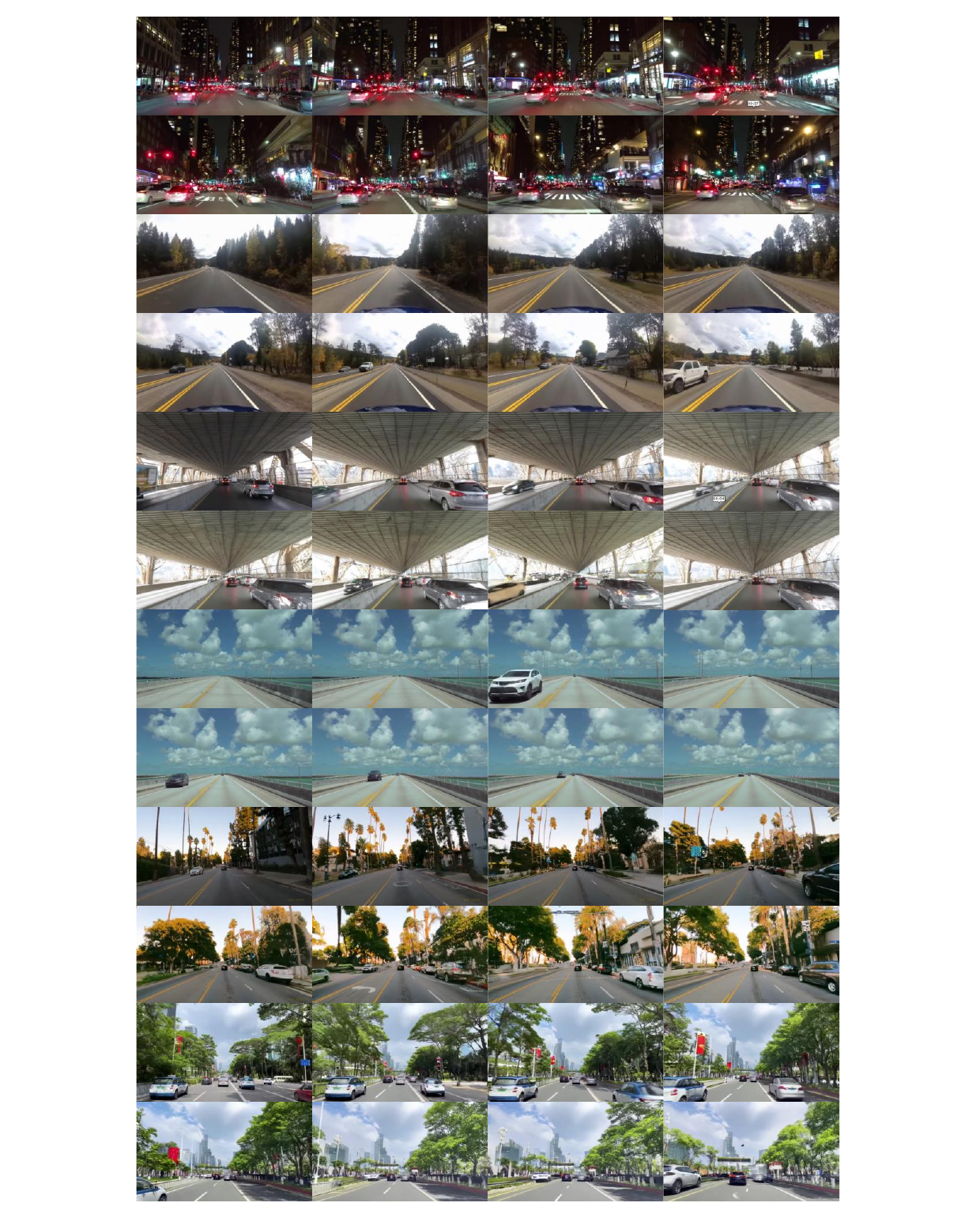}
   \caption{Generalization ability of \ourmethod\ in more scenarios.}
   \label{fig:vis_long}
\end{figure*}

\noindent\textbf{Diverse scenes}
In Fig.\ref{fig:vis_long}, we present the extended generation results across various scenes, demonstrating the robust generalization capability of our approach.

\begin{figure*}[h]
  \centering
   \includegraphics[width=1\linewidth]{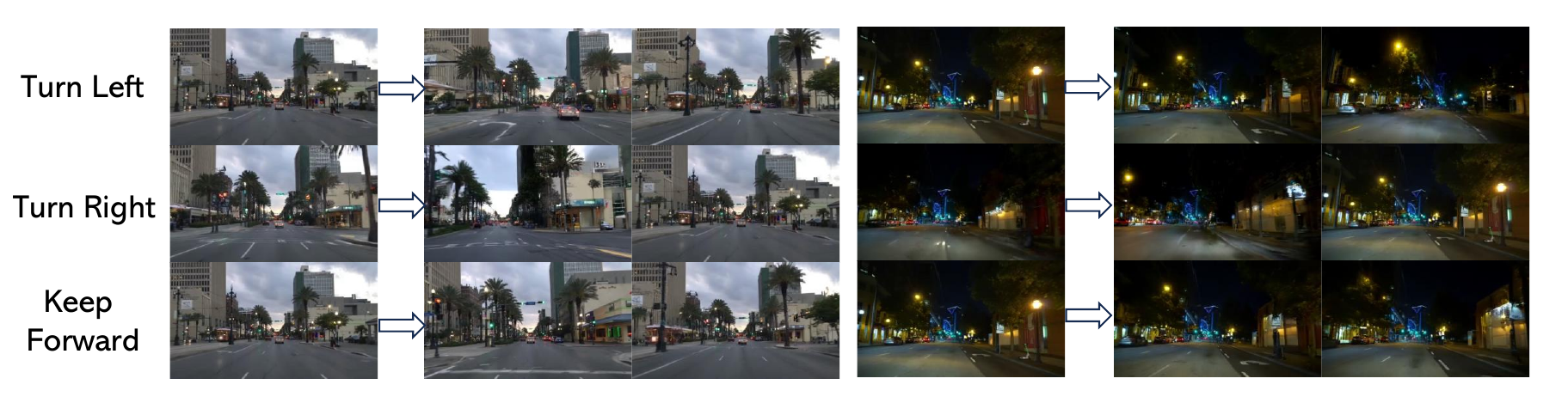}
   \caption{Action control ability of \ourmethod.}
   \label{fig:vis_action}
\end{figure*}

\noindent\textbf{Action control}
In Figure \ref{fig:vis_action}, we illustrate the controllability of our method on the OpenDV-2K dataset, adhering to the action module in Vista.

\begin{figure*}[h]
  \centering
   \includegraphics[width=1\linewidth]{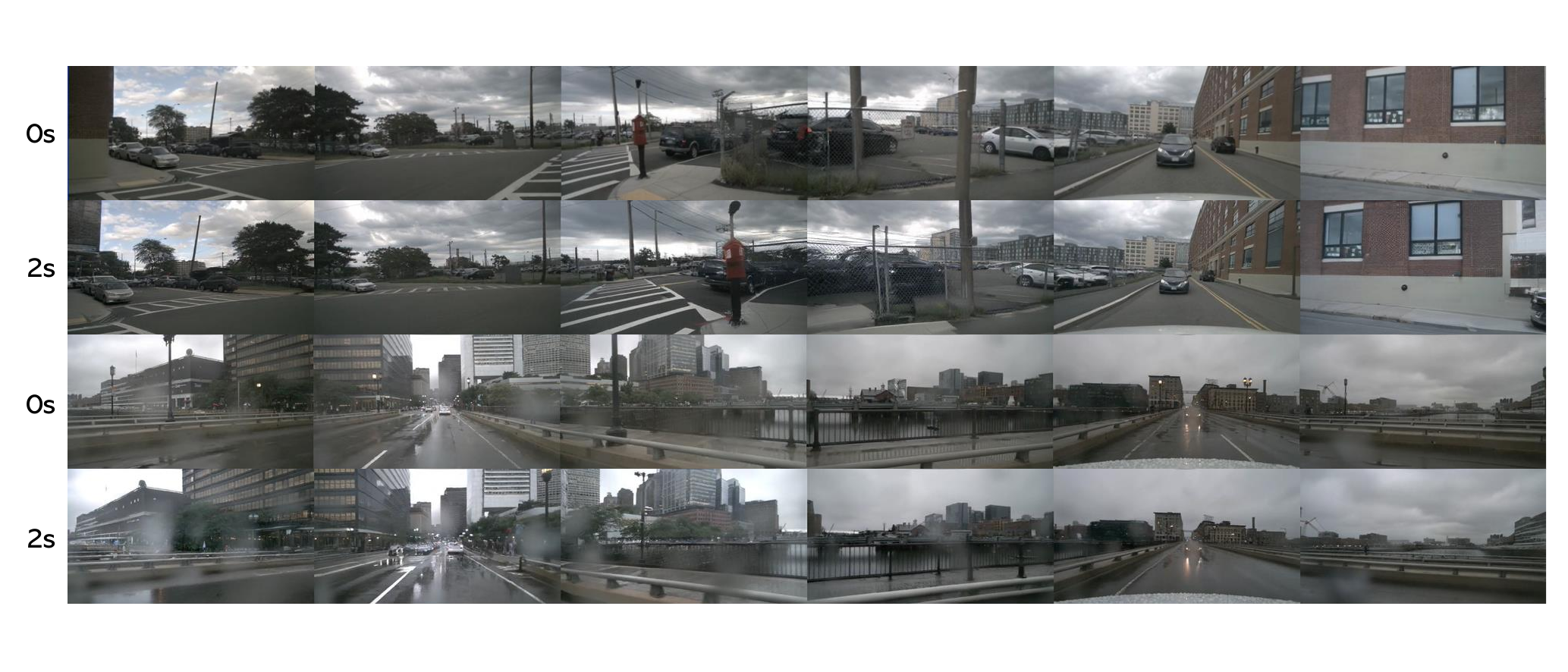}
   \caption{Generalization ability of multi-view videos.}
   \label{fig:vis_mview}
\end{figure*}

\noindent\textbf{Multi-view generation}
In Figure \ref{fig:vis_mview}, we show the multi-view generation ability coming from lifting our single-view model by extra view transformer blocks.

\section{Discussions}
\subsection{Differences to Vista}
Although both our method and Vista~\cite{vista} aim to construct a generalizable world model using the large-scale OpenDV-2K dataset, we highlight several key distinctions here. First, our findings suggest that relying solely on the Diffusion loss may not be optimal for building a world model. We introduce a complementary MR task, which has demonstrated robust generalization capabilities in representation learning tasks. Second, our model enables multi-view video generation through an additional training stage. This also illustrates that multi-view generation can benefit from a well-trained single-view model trained on a dataset encompassing significantly longer durations—over 1,700 hours in the OpenDV-2K dataset. Third, our model achieves longer prediction durations than Vista. As indicated by the slope of the metric changes in Fig.~\ref{fig:supp_abl_long}, our method maintains stable video prediction results, up to 15 seconds by autoregressive generation, whereas the generation quality of Vista degrades notably at this point. Moreover, we have found that our model can sustain stable generation over longer time periods across diverse scenes. Regarding quantitative evaluation, our model exhibits superior generalization capabilities, as evidenced by results on both the OpenDV-2K and Waymo datasets. On the standard nuScene benchmark, our approach also yields better results, with a 19\% improvement in FID and a 3\% decrease in FVD.

\subsection{Usage of Stable Diffusion 3}
Our baseline, built upon the SD3~\cite{sd3} model, yields superior results compared to GenAD (trained on SDXL~\cite{podell2023sdxl}) and performance slightly lower than Vista (initialized with SVD~\cite{svd}). Since both GenAD and our baseline are derived from image generation models, the improved performance of our baseline demonstrates the effectiveness of SD3. The superiority of SVD is attributed to its well-initialized temporal blocks, which have undergone multi-stage pre-training on extensive video datasets. Therefore, enhancing the data efficiency of SD3—as in our MR policy—and incorporating more video data present promising avenues to bridge this performance gap.

\subsection{Future impact of MR.}
In our method, MR acts as a complementary task to the diffusion loss, incorporating better video prediction abilities. Within the scope of representation learning, MR conducts context reasoning in a self-supervised way and can be generalized to various tasks. This aligns with our design: recovering the original MR at low noise levels using detailed local context. Our results show that diffusion models may excel in generating high-fidelity results but learn context reasoning slowly, which can be improved through the MR task. More generally, the effectiveness of MR shows that relying solely on diffusion may not be the optimal approach for driving world models. A similar inspiration can also be found in GAIA-1, where the prediction ability is decomposed into an auto-regressive model and a diffusion model. Exploring training targets for world models can be a promising direction.


\subsection{Limitations.}
Although better generalization ability and quality are achieved, there still exist some limitations that call for future works. (1) Controllability. Since we focused our main improvements on generalization ability and long-duration prediction, the action module follows the design of Vista. We have found several challenging cases in control, such as unreasonable commands. Similar to Vista, our method relies on resampling the nuScenes dataset to learn control ability. As a result, finding better feedback strategies and larger datasets for action learning is a promising direction. (2) Prediction of Uncertain Future. This phenomenon mainly arises when encountering complex traffic scenarios, especially when predicting the movement of each vehicle is difficult. (3) Generation of Non-Front View Images. Since multi-view capability is introduced only at the last training stage with a single nuScenes dataset, the images of non-front views lack exposure before this stage. Incorporating non-front view data at an earlier stage or adding more multi-view datasets (e.g., Waymo) may help address this problem.

%% file: tab/supp_abl_long.tex
\begin{figure*}[t]
  \centering
   \includegraphics[width=1\linewidth]{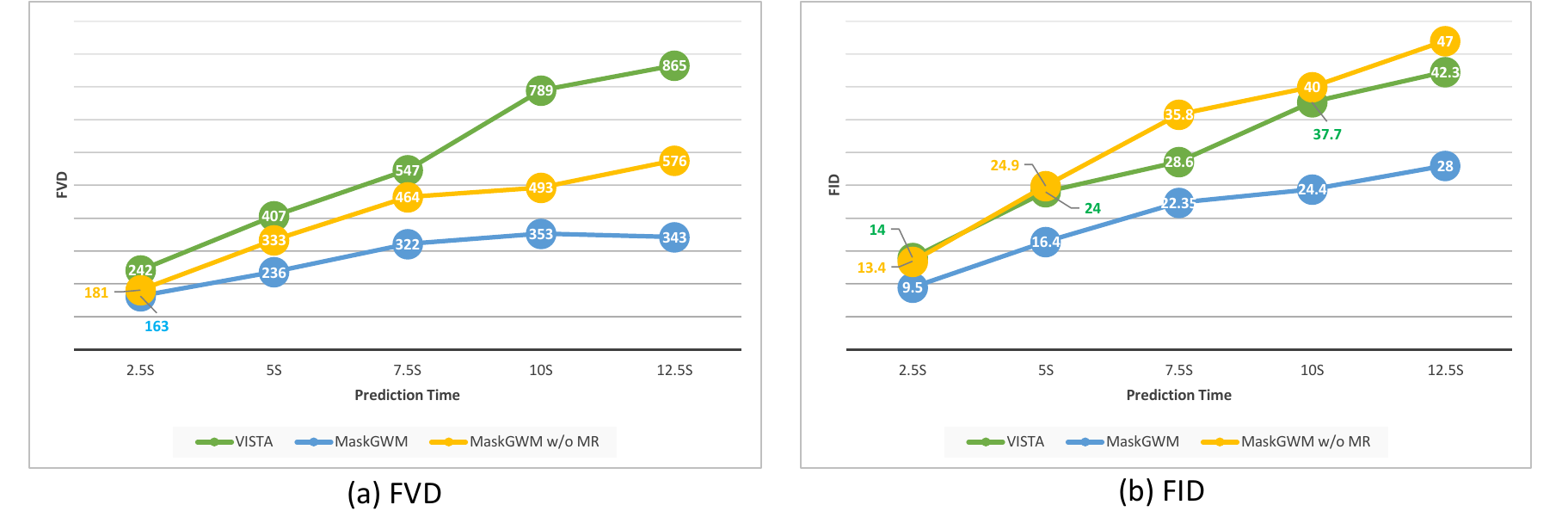}
   \caption{\textbf{Comparison of Long-horizon FVD metric on OpenDV-2K validation set.}  MR plays a crucial role in enhancing the capability to predict long video sequences, especially on FID.}
   \label{fig:supp_abl_long}
\end{figure*}

%% file: tab/supp_abl_mr.tex
\begin{table}[h]
    \centering
    \begin{tabular}{c|ccc}
    \toprule
    $r$ & $\mathcal{M}=\mathcal{M}_{spatial}$ & $\mathcal{M}=\mathcal{M}_{time}$ & $\mathcal{M}=\mathcal{\hat{M}}_{time}$\\
    \midrule
     0 & 136.5 & 136.5 & 136.5 \\
     0.1 & 125.8 & 133.2 & 123.7 \\
     0.25 & 116.7 & 142.6 & 121.3 \\
     0.4 & 155.9 & 179.1 & 149.8 \\

    \bottomrule
    \end{tabular}

    \caption{FVD comparisons on mask ratio. }
    \label{tab:supp_abl_mr}
\end{table}

%% file: tab/supp_abl_ns.tex
\begin{table}[h]
    \centering

    \begin{tabular}{cc|cc}
    \toprule
    row\&shift att. & $r$ & \textbf{FVD$\downarrow$} & \textbf{FID$\downarrow$} \\
    \midrule
     & 0 & 107.2 & 7.5 \\
    \midrule
     $\surd$ & 0.25 & 92.5 & 5.6 \\

    \bottomrule
    \end{tabular}

    \caption{Ablations on nuScene dataset. }
    \label{tab:supp_abl_ns}
\end{table}